\documentclass{article}
\usepackage{spconf,amsmath,epsfig}
\usepackage{xcolor, soul}
\usepackage{siunitx}
\usepackage{booktabs}
\usepackage{tabularx}
\usepackage{hyperref}

\sethlcolor{yellow}

\title{Landslide mapping from Sentinel-2 imagery through change detection}
\threeauthors
  {Tommaso Monopoli}
  {LINKS Foundation\\	AI, Data \& Space\\ Turin, Italy}
  {Fabio Montello}
  {LINKS Foundation\\	AI, Data \& Space\\ Turin, Italy}
  {Claudio Rossi}
  {LINKS Foundation\\	AI, Data \& Space\\ Turin, Italy}

\begin{document}
%
\maketitle
\def\thefootnote{}\footnotetext{This work was partially funded by the Horizon Europe projects ATLANTIS (GA n.101073909) and RescueME (GA n.101094978), and by the project NODES through the MUR—M4C2 1.5 of PNRR under Grant ECS00000036.}

\begin{abstract}
Landslides are one of the most critical and destructive geohazards.
Widespread development of human activities and settlements combined with the effects of climate change on weather are resulting in a high increase in the frequency and destructive power of landslides, making them a major threat to human life and the economy.
In this paper, we explore methodologies to map newly-occurred landslides using Sentinel-2 imagery automatically. All approaches presented are framed as a bi-temporal change detection problem, requiring only a pair of Sentinel-2 images, taken respectively before and after a landslide-triggering event.
Furthermore, we introduce a novel deep learning architecture for fusing Sentinel-2 bi-temporal image pairs with Digital Elevation Model (DEM) data, showcasing its promising performances w.r.t. other change detection models in the literature.
As a parallel task, we address limitations in existing datasets by creating a novel geodatabase, which includes manually validated open-access landslide inventories over heterogeneous ecoregions of the world. 
We release both code and dataset with an open-source license. 
\end{abstract}

\section{Introduction}
\label{sec:introduction}

Landslides are one of the most common catastrophic geological events, causing extensive economic losses, posing a serious threat to human settlements, and often resulting in fatalities \cite{schuster1986economic}.
The damage that landslides cause to life and property has been ever-increasing over recent years, mainly because of the wider spread of human activities and urban development in areas that are more prone to landslides. Moreover, the most recent IPCC report estimates that by the end of this century, we should expect twice as intense precipitation events compared to today as a result of climate change, which will certainly increase the frequency and destructive power of landslides \cite{ipcc23} \cite{li2020influence}.
It is widely recognized that rapid landslide mapping is crucial for effective disaster response and damage mitigation, especially in the immediate aftermath of such catastrophic events \cite{galli2008comparing}. The timely discovery and delineation of areas impacted by landslides can significantly enhance the overall disaster management process, enabling authorities to make informed decisions, allocate resources efficiently, and estimate the impact on affected communities and infrastructures.

In this work, we explore deep learning methodologies for the automatic mapping of landslides from Sentinel-2 images, framing the problem as a change detection approach. The main contributions of this work are (i) the development of a novel bitemporal-bimodal deep learning architecture for change detection and (ii) the creation of a globally diverse geodatabase encompassing and harmonizing several manually validated open-access landslide inventories.

\begin{figure*}[t]
    \centering
    \includegraphics[width=\textwidth]{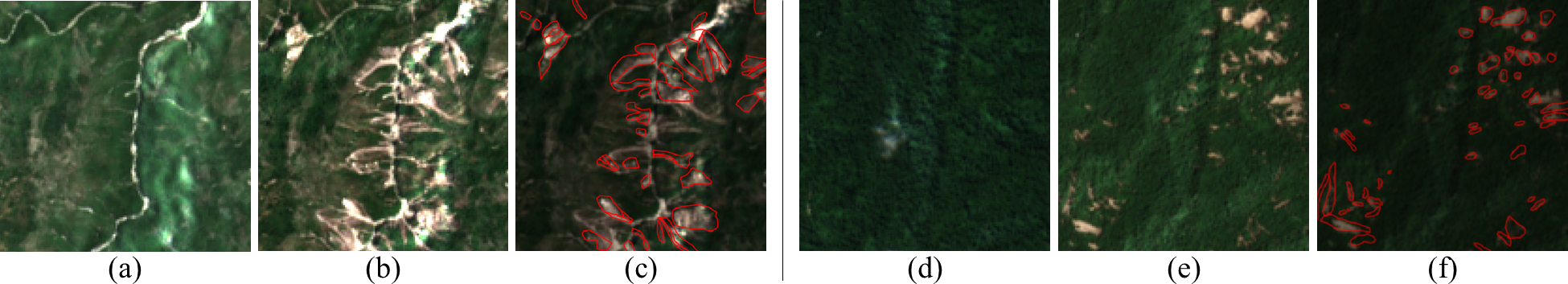}
    \caption{Sentinel-2 images before and after a landslide-triggering catastrophic event. (a)-(b)-(c): a sample from Haiti inventory. (d)-(e)-(f): a sample from Indonesia inventory. (c) and (f) show landslide polygons which are present in the respective inventories overlayed on the post image (dimmed for visibility).}
\label{fig:prepost}
\end{figure*}

\section{Related Works}
\label{sec:related_works}
Some recent research works have explored innovative approaches for automatic landslide mapping based on machine learning and deep learning techniques \cite{tehrani2022machine} \cite{sreelakshmi2022landslide} \cite{ma2021machine}. The wider availability of landslide inventories and commercial and non-commercial satellite data has fostered the creation of supervised machine learning methodologies for the landslide delineation task.

However, despite the promising results showcased, existing works have several limitations:
\begin{itemize}
    \item Most previous studies focus on a single ecoregion, i.e., a geographical area exhibiting homogeneous properties in terms of lithology, morphology, landforms, soil composition, vegetation type, etc. When considering a supervised machine learning approach, the limited geographical scale of the training data inherently hampers the model's ability to generalize to new, visually different areas \cite{nagendra2022constructing}. Therefore, effort is needed to create an extensive database of landslide events, with a focus on heterogeneity of ecoregions, landslide sizes and landslide triggering causes.
    \item Most works deal with commercial (very) high-resolution data (e.g., \cite{ghorbanzadeh2019evaluation, yu2021matrix, su2021deep}). While high-resolution imagery can capture fine-grained details that might be useful to delineate even small landslides and improve segmentation performances, acquiring the data is relatively expensive. Moreover, high-resolution satellites typically capture less ground area with each flyover due to their narrow swath width \cite{cai2023learning}. Both facts make using high-resolution satellite imagery for emergency response use cases impractical, especially when the region of interest has a large area of several km$^2$. Using medium-resolution non-commercial satellite data, such as Sentinel-2 optical imagery, can be a solution, but it is still not as widely explored in the literature \cite{ghorbanzadeh2021a}.
    \item Existing landslide segmentation models typically do not work with bitemporal pairs of images (taken before and after a landslide-triggering event), but rather with a single image acquired after the landslide event. This leads to the detection of both newly activated landslides, associated with the event of interest, and old landslides, which may have happened in the past. This ambiguity is not desirable in emergency response scenarios, in which it is critical to rapidly map the impact area of newly activated events. Therefore, we propose to frame the landslide delineation as a change detection task.
\end{itemize}

\section{Materials and methods}
\label{sec:materials_and_methods}

\subsection{Database creation}
\label{sec:database_creation}
To tackle the visual homogeneity problem, we harmonize and combine several recent open-access landslide inventories from several ecoregions of the world, constructing a global and diverse landslide database of manually-validated landslide polygons. The database encompasses landslides of different sizes (very small, $\sim$100 m$^2$, to very large $\sim$100,000 m$^2$) and triggered by different catastrophic events (earthquakes, heavy rainfall). So far, our database includes a total of 34,920 distinct landslide polygons.
Figure \ref{fig:prepost} depicts pairs of Sentinel-2 images taken before and after the occurrence of landslides in the considered inventories.
A comprehensive list of open-access landslide inventories included in our database is reported in Table \ref{tab:inventories}.

\begin{table*}[t]
\centering
\caption{Open-access landslide inventories included in our database}
\label{tab:inventories}
\begin{tabularx}{\textwidth}{cccX}
\toprule
Inventory  &  References  &  N. polygons  &  Landslide-triggering event(s)\\
\midrule
Tiburon peninsula (Haiti) & \cite{inv:haiti1} & 14,482 & M7.2 earthquake which striked on 2021-08-14, followed by Hurricane Grace, which brought heavy rainfall on 2021-08-17 \\
\midrule
Central Sulawesi (Indonesia) & \cite{inv:palu1, inv:palu2} &  6,749  & M7.5 supershear earthquake occurred on 2018-09-28 \\
\midrule
Luding county (China) & \cite{inv:luding1, inv:luding2, inv:luding3} & 8,194 & M6.8 earthquake occurred on 2022-09-05 \\
\midrule
Mesetas (Colombia) & \cite{inv:mesetas1} & 838 & M5.7 earthquake occurred on 2019-12-24 \\
\midrule
Iburi region (Japan) & \cite{inv:iburi1, inv:iburi2} & 4,657 & M6.6 earthquake occurred on 2018-09-06 \\
\bottomrule
\end{tabularx}
\end{table*}

Each inventory in the database is associated with the date of the occurrence of the natural disaster which triggered the landslides in the region. We make the simplifying assumption that all the landslides mapped in each inventory co-occurred in that date, so that any other date can be classified as either "pre" or "post" event. For the Haiti inventory, we assume that all landslides occurred between 2021-08-14 and 2021-08-17.

\subsection{Dataset download and processing}
\label{sec:data_processing}
For each region of interest of each inventory in the database, pre and post event Sentinel-2 L2A images\footnote{\url{https://planetarycomputer.microsoft.com/dataset/sentinel-2-l2a}} are acquired, roughly ranging between 3 months before and 1 month after the event. All 12 Sentinel-2 L2A bands are retrieved: B01, B02, B03, B04, B05, B06, B07, B08, B8A, B9, B11, B12. The spatial resolution of these bands ranges from 10 to 60 m/px. We manually discard S2 images acquired in the winter if the region is covered in snow on the date of acquisition, since snow can potentially cover landslide scars on the ground. Due to the extended presence of clouds in the images, we use a cloud detector trained on the CloudSen12 dataset \cite{cloudsen12L2A} to generate a cloud mask. Specifically, the detector segments each image into four categories: clear sky, thick clouds, thin clouds, cloud shadow.

Moreover, for each region of interest we also retrieved the ALOS-PALSAR Digital Elevation Model (DEM) data\footnote{\url{https://planetarycomputer.microsoft.com/dataset/alos-dem}}, characterized by a spatial resolution of 30m and generated from remote sensor data acquired between 2006 and 2011. To enrich the data, we compute additional products from DEM data, namely the terrain slope (degrees in $[\ang{0},\ang{90}]$) and aspect direction (in degrees). Since aspect is a circular variable, which may be difficult for the deep learning model to deal with, we encode it into two distinct bands, respectively the sine and cosine of the aspect angle. Each DEM product will then be a stack of 4 bands: elevation, slope, aspect (sin), aspect (cos).

Finally, by burning polygons of each inventory into a raster image with spatial resolution of 10 m/px, a ground-truth segmentation mask is obtained for each region.

In order to train a deep learning segmentation model, some data preprocessing steps are applied. First, all bands of the S2 images and DEM data are upscaled to a common resolution of 10 m/px. Then, we consider all the possible S2 pre-post pairings for each region. We tile each S2 image pair and its corresponding ground-truth mask and DEM data by extracting $256 \times 256$ patches (which at a resolution of 10m/px corresponds to an area of roughly 6.55 km$^2$), overlapping with stride 128. Then, to filter out any patches with too low informative content, we only keep those pairs of S2 patches which meet the following requirements:
\begin{itemize}
    \item their ground truth mask contains at least one “visible” landslide pixel (i.e. not covered by clouds)
    \item the union of the predicted thick and thin cloud cover masks for the pre and post patches (i.e. the "merged" cloud cover associated to the pair) covers at most 20\% of the patch area
    \item all pixels are valid (i.e. not marked as S2 NODATA value)
\end{itemize}

Finally, S2 pixel values are divided by 10,000 and clipped to $[0,1]$, consistently with SSL4EO-S12 dataset \cite{ssl4eo} data pre-processing pipeline. Elevation values in DEM data are divided by 5,000 and slope angles are normalized to $[0,1]$.


\subsection{Data augmentations}
\label{sec:data_augs}
To enhance the visual heterogeneity of the dataset and increase the generalization capability of the models, we apply several data augmentation techniques to the patches at training time. These include geometric transformations (applied both to S2, DEM and ground truth mask data) such as random horizontal/vertical flips and random rotations, and color transformations (applied only to S2 patches) such as brightness and contrast jittering and histogram matching.

In particular, band-wise histogram matching can be considered as a synthetic variant of seasonal contrast \cite{seasonalcontrast}: by matching the band histograms of an S2 image to those of another S2 image from a possibly different season, the resulting image teaches the model to be invariant to seasonal changes, while retaining essential semantic visual information and correlation between pixels.

\section{Experiments and results}
\label{sec:experiments}
We frame the landslide mapping task as a change detection problem: delineating landslides which are present in a post-event S2 image but not in a pre-event one. We train several state-of-the-art change detection models: Unet-Siam-Diff \cite{unet_siam} with ResNet50 encoder, BIT \cite{bit}, SEIFNet \cite{seifnet}, TinyCD \cite{tinycd}.
These bitemporal models only support pre-post image pairs, therefore DEM data cannot be used.

\begin{figure*}[t]
    \centering
    \includegraphics[width=0.96\textwidth]{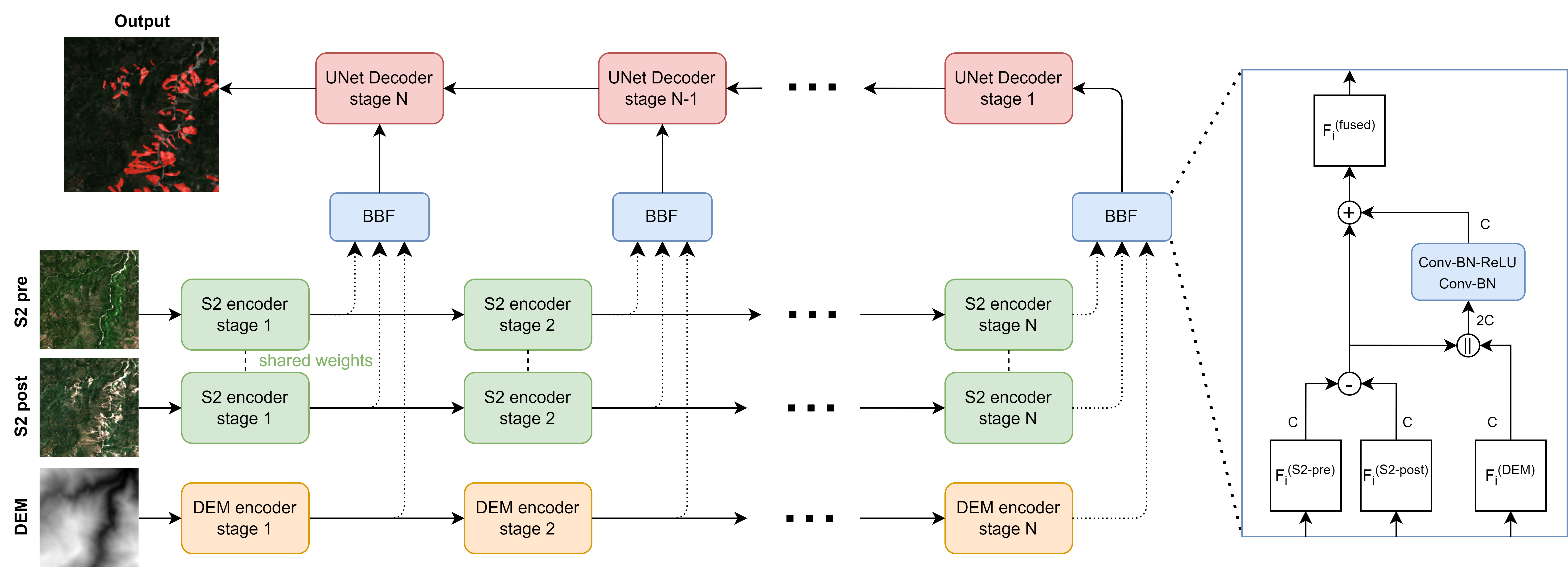}
    \caption{BBUnet architecture, with detail on the Bitemporal-Bimodal Fusion (BBF) module.}
\label{fig:bbunet}
\end{figure*}

For this reason, we introduce a novel bitemporal and bimodal change detection model which can handle both a bitemporal pair of remote sensing images (e.g., S2 image pair) and an additional, single-temporal data stack (e.g., DEM data) which enriches the bitemporal pair with potentially useful additional contextual information. Its architecture is a modified Unet-Siam-Diff, with the addition of a simple convolutional fusion module inspired by the one introduced in \cite{lu2023dual}. We refer to this model as BBUnet (standing for Bitemporal-Bimodal Unet). The model's architecture is depicted in Figure \ref{fig:bbunet}.

We split our dataset into a training set (with all bitemporal pairs generated from Central Sulawesi, Luding county, Mesetas and Iburi region inventories), a validation and test set (each having half the pairs generated from Haiti inventory). Patches with less than 200 visible landslide pixels are excluded from the validation and test set, to focus inference only on patches with higher landslide presence. The number of samples in the training/validation/test sets are respectively 55,602 / 1,102 / 1,102. We train each model for 50 epochs, with AdamW optimizer, batch size of 64, learning rate of 0.01 with exponential annealing, weight decay of 0.0001.
The loss function is a weighted binary cross-entropy loss with weight of the "landslide" class equal to 5, in order to tackle its imbalance w.r.t. background class.
Pixels predicted as clouds in the merged cloud cover mask are ignored (i.e. excluded from loss computation). To obtain the final segmentation mask, the output logits are rectified through the sigmoid function and thresholded at 0.5. The model's weights after each epoch are stored; as the training ends, the model checkpoint achieving the lowest loss on the validation set is taken as final and tested on the test set.

Results on the test set for each final model are reported in Table \ref{tab:results}.

\begin{table}[t]
\centering
\caption{Performance metrics of the models on the test set}
\label{tab:results}
\begin{tabularx}{\linewidth}{cXXX}
\toprule
Model  &  F1 score  &  Precision  &  Recall \\
\midrule
BIT  &  0.306  &  0.497  &  0.221 \\
SEIFNet  &  0.299  &  0.576  &  0.201 \\
TinyCD  &  0.257  &  0.580  &  0.165 \\
UNet-Siam-Diff  &  0.341  &  0.525  &  \textbf{0.252} \\
\midrule
BBUnet (ours) &  \textbf{0.348}  &  \textbf{0.582}  &  0.249 \\
\bottomrule
\end{tabularx}
\end{table}

As can be noticed, achieved performances are generally not so high. Despite all efforts put into the dataset construction and choice of data augmentations, we argue that the not so perfect quality of the manually-mapped landslide inventories and the relatively low resolution of S2 satellite in comparison to landslides' size constitute an intrinsic problem, as well as the major challenge of this research work.
Nonetheless, achieved precision values are generally high, meaning that the models effectively learn to correctly characterize and delineate landslides without too much false positives.

As reported, BBUnet achieves the highest F1 score among all experimented models. Its recall is almost on par with the highest one, achieved by UNet-Siam-Diff, and its precision is on par with TinyCD (which however achieves an unacceptably low recall). These results suggest that the addition of DEM data to support Sentinel-2 images is beneficial to the landslide delineation task. This could be expected, as there is a proven high correlation between the occurrence of landslides and terrain features, such as its slope and aspect \cite{youssef2023landslide}.

\section{Conclusion}
\label{sec:conclusion}
In this work, we introduced a large geodatabase of landslide events encompassing several manually-validated landslide inventories, with a focus on heterogeneity of ecoregions and landslide trigger factors.
To the best of our knowledge, our geodatabase is the largest and most heterogeneous collection of landslides events available in open access.

Moreover, we introduced a deep learning methodology to delineate landslide events, framed as a change detection task on Sentinel-2 pre- and post- event acquisitions, and studied its performances in various experimental settings.
To this end, we introduced a novel bitemporal and bimodal change detection architecture, BBUnet, capable of fusing DEM data with Sentinel-2 bitemporal image pairs. Its improved performances in comparison to other change detection models highlight that landslide delineation from low resolution satellite imagery can benefit from additional contextual information such as pre-event terrain slope and aspect.

To foster future research and scientific cooperation on the topic of landslide segmentation, we share our landslide database and code repository at \url{https://github.com/links-ads/igarss-landslide-delineation}.

While our findings are promising, we acknowledge that several aspects of our research remain open, including meticulous data quality assessment for the existing landslide inventories, how to handle incorrectly mapped/missing landslides in an inventory and how to cope with Sentinel-2 relatively low resolution in comparison to a landslide's average size. These challenges shall be addressed in future work, along with improvements in the proposed methodology.

\bibliographystyle{IEEEbib}
\bibliography{refs}

\end{document}